\documentclass[
]{ceurart}

\usepackage{listings}
\usepackage{latexsym}
\usepackage{amssymb}
\usepackage{amsmath}
\usepackage{amsthm}
\usepackage{booktabs}
\usepackage{enumitem}
\usepackage{graphicx}
\usepackage{color}
\usepackage{makecell}
\usepackage{xcolor,colortbl}
\usepackage{array,multirow}
\usepackage{comment}
\usepackage{url}
\usepackage{caption}
\usepackage{subcaption}
\usepackage{siunitx}
\usepackage{rotating}
\usepackage{epigraph}
\usepackage{arydshln}

\begin{document}

\copyrightyear{2025}
\copyrightclause{Copyright for this paper by its authors.
  Use permitted under Creative Commons License Attribution 4.0
  International (CC BY 4.0).}

\conference{Identity-Aware AI workshop at 28th European Conference on Artificial Intelligence,
  October 25, 2025, Bologna, Italy}

\title{On the Interplay between Musical Preferences and Personality through the Lens of Language}


\author[]{Eliran Shem-Tov}[%
email=eliran.shemtov@gmail.com,
]
\cormark[1]
\address[]{The Academic College of Tel Aviv--Yaffo, Tel Aviv--Yaffo, Israel}

\author[]{Ella Rabinovich}[%
email=ellara@mta.ac.il,
]


\begin{abstract}
Music serves as a powerful reflection of individual identity, often aligning with deeper psychological traits. Prior research has established correlations between musical preferences and personality, while separate studies have demonstrated that personality is detectable through linguistic analysis. Our study bridges these two research domains by investigating whether individuals' musical preferences leave traces in their spontaneous language through the lens of the Big Five personality traits (Openness, Conscientiousness, Extroversion, Agreeableness, and Neuroticism). Using a carefully curated dataset of over 500,000 text samples from nearly 5,000 authors with reliably identified musical preferences, we build advanced models to assess personality characteristics. Our results reveal significant personality differences across fans of five musical genres. We release resources for future research at the intersection of computational linguistics, music psychology and personality analysis.

\end{abstract}

\begin{keywords}
  natural language processing \sep
  personality traits \sep
  music psychology
\end{keywords}

\maketitle


\section{Introduction}
\label{sec:introduction}

\epigraph{\textit{"Our language is the reflection of ourselves. A language is an exact reflection of the character and growth of its speakers."}} {{Cesar Chavez (civil rights activist)}}


Music is a powerful expression of individuality, often reflecting deeper aspects of one's character and personality. The relationship between musical preferences and personality has long been a subject of interest in psychology \citep{cattell1953measurement, rentfrow2003re, bonneville2013music, ferwerda2017personality}. Prior studies indicate that certain genres are associated with specific personality traits, suggesting that music genre preferences offer valuable insights into individuals' psychological profile \citep{bonneville2013music, nave2018musical, anderson2021just}.

Originally introduced by \citet{goldberg1992development}, the Big Five personality traits --- \textbf{O}penness (OPN), \textbf{C}onscientiousness (CON), \textbf{E}xtroversion (EXT), \textbf{A}greeableness (AGR), and \textbf{N}euroticism (NEU), collectively denoted by the acronym "OCEAN" --- were established in psychology as one of the most common ways for assessing one's personality. Over decade of computational studies have shown that our personality is reflected in our language (to the extent detectable by automatic tools), driving the development of methods for personality assessment in linguistic productions \citep{greenberg2015musical, sun2018personality, peters2024large}.

Our study aims to connect the two lines of research -- we explore the \textit{relationship between musical preferences and personality traits, as manifested in authentic written language}. 
Specifically, we ask if there are detectable (similarities and) differences in the Big Five personality traits in the language of people with various musical preferences, when authoring spontaneous textual content on social media. We consider five popular and diverse musical genres: Classical music, Hip-Hop, Metal, Indie and Electronic. The hypothesis driving this research posits that despite the inherent complexity of automatic personality analysis, the differences in personality traits of people with various musical preferences "shine through" their language, to the extent that can be automatically captured through advanced natural language processing techniques.

We explore the hypothesis on a carefully curated large and diverse dataset, consisting of over 500,000 text samples written by nearly 5,000 authors with reliably identified musical preferences. These samples were collected from a range of public non-music-related online forums, ensuring a broad topical representation. Each participant was linked to a single dominant music genre they frequently engage with, enabling unbiased textual personality extraction and a focused analysis of the relationship between personality traits and musical preferences. Leveraging text generation capabilities of contemporary LLMs we then gather high-quality data for training personality detectors along the five dimensions: Openness, Conscientiousness, Extroversion, Agreeableness, and Neuroticism. 

The models were further applied to the text samples authored by fans of different music genres. Statistical tests reveal significant and reliably detected differences in personality characteristics of individuals with diverse musical preferences. As an example, Classical music enthusiasts show \textit{higher levels of Agreeableness} and \textit{lower levels of Extroversion}, while Hip-Hop fans have \textit{lower Agreeableness} and \textit{higher Neuroticism}. Additionally, Metal fans exhibit a \textit{higher propensity for Neuroticism}, and Electronic music listeners show \textit{greater Openness}.

Our contribution in this study is, therefore, manyfold: First, we release a carefully collected dataset of authentic linguistic productions of users with diverse and reliably identified musical preferences. Second, we train models for accurate personality detection from text, and release their training data. Finally, through large-scale empirical analysis, we shed a new and interesting light on the association between musical preferences and personality traits, as manifested in our language. All our data and code are available at \url{https://github.com/eliranshemtov/Musical-Preferences-And-Textual-Expression}.
\section{Related Work}
\label{sec:related-work}

\paragraph{Personality and Musical Preferences}
Although sparse, first prior art goes back to 1950s and reveals clear ties between musical preferences and personality traits. These studies typically rely on \textbf{self-reported} personality traits (primarily OCEAN), and are of relatively small-scale. \citet{cattell1953measurement} and \citet{cattell1954musical} suggested that music could satisfy deep and unconscious needs, thus providing insights into personality. Specifically, they developed the IPAT Music Preference Test, which identified stable music-preference factors reflecting unconscious aspects of personality. \citet{rentfrow2003re} found that Openness to experience correlated with a preference for Classical music, while Extroversion was linked to energetic and rhythmic music. Moreover, Agreeableness was associated with upbeat and conventional music. 

Additional studies conformed with earlier works and found that the Openness to experience trait is correlated with many music genres, including new age, classical, world, blues, country, folk, jazz, and alternative \citep{bonneville2013music, ferwerda2017personality}. Furthermore, population scored high in Conscientiousness showed a negative correlation with folk and alternative music, while Extroversion correlated with R\&B (Rhythm and Blues) and Rap music. Agreeableness positively correlated with Country and Folk, and Neuroticism only showed a positive correlation with Alternative music \citep{ferwerda2017personality}. A recent study by \citet{greenberg2022universals} examined musical preferences across 53 countries, revealing that Extroversion was correlated with stronger reactions to contemporary styles, Openness was correlated with "sophisticated music", whereas Neuroticism was connected to intense musical styles, reflecting "inner angst and frustration".


\paragraph{Personality Detection from Text} Automatic detection of personality traits from text roots back to the pioneering work of Francis Galton in 1884 (reprinted in 1949 -- \citet{galton1949measurement}). The author claimed that personality could be effectively captured through the adjectives found in language and written text, laying the groundwork for future studies in personality extraction. Since then, multiple (early) works used the combination of linguistic cues and automatic language processing methods for personality detection, focusing on the Big Five traits extracted from the Essays dataset \citep{pennebaker1996cognitive, pennebaker1999linguistic, mairesse2007using}.

The release of the MyPersonality dataset by \citet{kosinski2015facebook} --- not a non-controversial outcome of data collection through a Facebook application --- sparked another line of research, using more advanced methods, inspired by a range of deep-learning architectures \citep{majumder2017deep, sun2018personality, zheng2019predicting, ren2021sentiment, yang2023getting}. Working with MyPersonality \citep{kosinski2015facebook}, whose support was officially discontinued in 2018, and Essays \citep{pennebaker1999linguistic}, researches achieved higher accuracy: around 0.70 on the binary (high or low) personality trait detection task. 

Advancements in pre-trained large language models (LLMs) put forward additional opportunities for the field of personality detection from text. Approaches vary from transfer learning for classification \citep{yang2023getting, alshouha2024personality}, through graph-based methods \citep{yang2021psycholinguistic, zhu2022contrastive, zhu2024data}, to using the most powerful SOTA models (e.g., GPT4) in zero-shot scenario \citep{peters2024large}. Notably, the studies still manage to achieve a moderate accuracy on the two datasets, with GPT4 obtaining the very weak correlation (Pearson's $r${=}0.31) to self-reported scores in the MyPersonality dataset.

\paragraph{Personality, Musical Preferences and Language}
Our study is the first, to the best of our knowledge, to explore the association of personality traits detected from text, to musical preferences.

\section{Dataset}
\label{sec:dataset}

We collected a large and diverse dataset of spontaneous and authentic written content from individuals with distinct musical preferences. We aimed to gather texts that reflect a wide range of commonly discussed topics, deliberately excluding any direct references to music, to avoid potential confounds and ensure that the analysis accurately captures the relationship between musical preferences of an individual and their personality traits.

\subsection{Users with Various Music Preferences}
Our dataset in this study was collected from the Reddit discussion platform:\footnote{\url{https://www.reddit.com/}} a well-known, highly organized, and topic-categorized home of over 2.3 billion monthly active users,\footnote{\url{https://arc.net/l/quote/byjckxas}} in over 130 thousand communities, also known as subreddits. Reddit maintains various active communities (subreddits) for discussions of music in different genres. Among the most active musical subreddits are Indie, Electronic, Hip-Hop, Metal, and Classical music. Offering five distinct preferences, these genres' fans are the main subject of our study.

First we retrieved $\sim$1M most recent posts from each of the threads, and identified top-K most active users in each subreddit, where K was set to 1,000, excluding those in the "intersection" with other genres: users that authored posts or comments in any of the other four music communities. As a concrete example, the most recent 1,000,154 comments in the Metal subreddit (\texttt{r/metal}) were posted by 74,055 Redditors, where 1,000 most active authors (w/o any activity in the other subreddits), were considered for this study. We refer to the set of \textit{active} 1,000 users associated with a \textit{single genre} (e.g., Metal) as reliable enough for the purpose of this work. Similar assumptions about the association between Reddit authors' activity and their meta properties were shown reliable for the country of origin \citep{rabinovich2018native}.

\subsection{Dataset Collection}
Given a Reddit \texttt{UserID} of each of the 5,000 authors, the entire user’s digital footprint can be retrieved from the platform, providing direct access to a person’s linguistic productions across a variety of topics, spanning several years. We applied multiple filtering and cleanup steps: (1) excluded subreddits (in)directly related to music,\footnote{About 150 threads were excluded per manual inspection.} (2) aiming at long enough, ideally paragraph-length texts for personality analysis, we filtered out texts shorter than 40 tokens,\footnote{The minimal number of tokens is aligned with the mean paragraph length in our training data (see Section~\ref{sec:methodology}).} and (3) removed duplicate entries. These steps resulted in a clean and high quality dataset; the dataset statistics are reported in Table~\ref{tbl:dataset-stats}. 

\begin{table}[h!]
\caption{Datasets statistics: the total number of texts (posts and comments) collected for each musical genre, total and mean number of texts per user. Note that we ended up with slightly less than 1K users per genre since some did not have any texts satisfying the min length threshold. 
}
\centering
\begin{tabular}{lrrr}
\toprule
genre & users & total texts & mean texts per user \\
\midrule
{Classical} & 982 & 170,251 & 173.37 \\
{Hip-Hop} & 993 & 121,538 & 122.39 \\
{Electronic} & 886 & 97,063 & 109.55 \\
{Indie} & 871 & 84,314 & 96.80 \\
{Metal} & 951 & 102,650 & 107.94 \\
\midrule
total & 4,683 & 575,816 & -- \\
\bottomrule
\end{tabular}
\label{tbl:dataset-stats}
\end{table}

\begin{figure}[h!]
\centering
\includegraphics[width=0.90\linewidth]{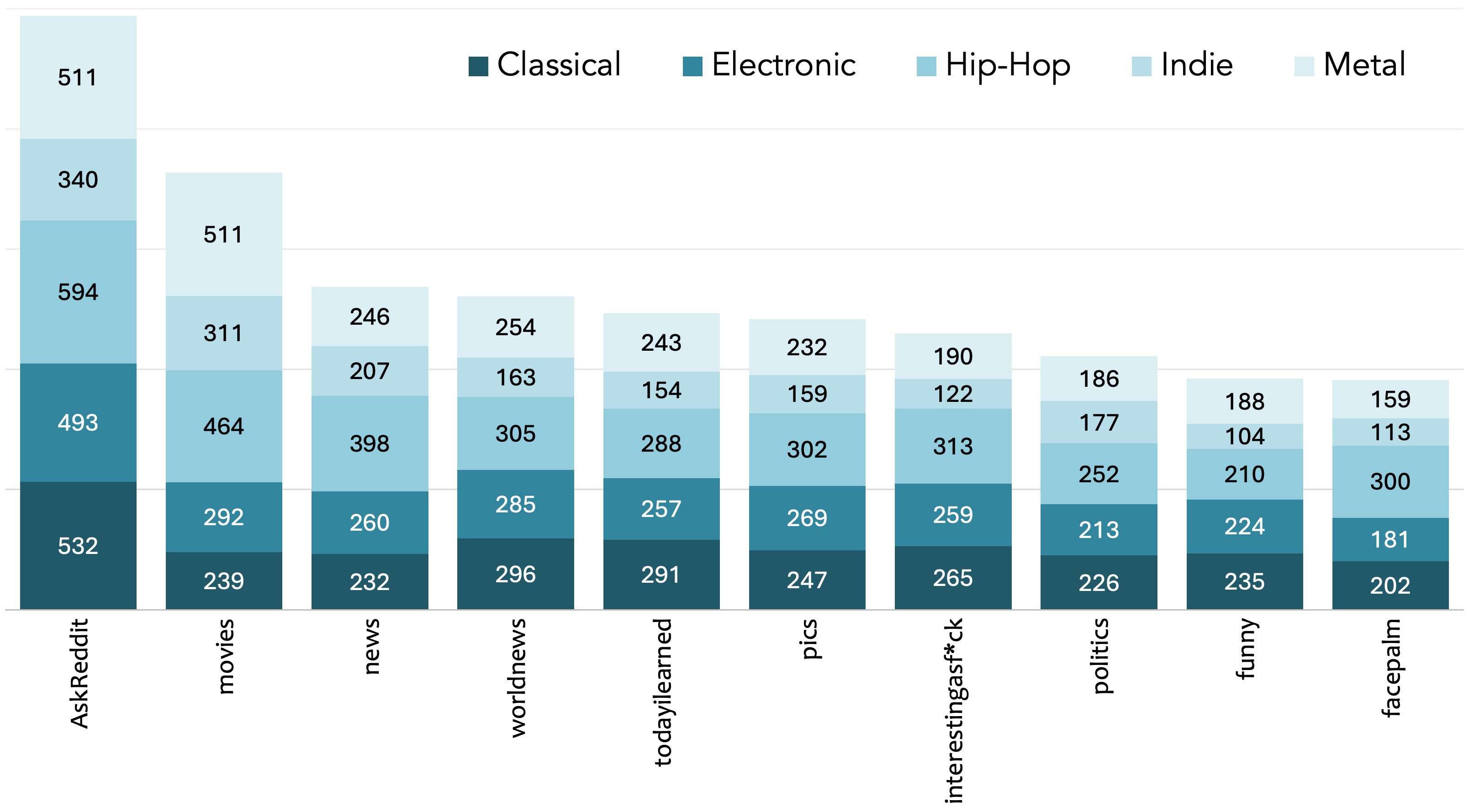}
\caption{Distribution of the number of users (out of the pool) that posted or commented at least once in one of the top-10 most popular subreddits in the data, by genre. The various topical threads are represented roughly equally by the fans of the five musical genres.
}
\label{fig:subreddit-dist}
\end{figure}

Additionally, Figure~\ref{fig:subreddit-dist} reports the distribution of user participation in the top-10 most-popular communities in our collected data, split by musical preference. Notably, the various topical threads are represented roughly equally by the fans of the five musical genres in our study, suggestive of a dataset free of topical confounds. 

\section{Personality Detection from Language: Datasets and Approaches}
\label{sec:personality-detection}


A typical approach to personality detection from text includes the analysis of written or spoken language to identify and predict various personality traits of individuals; these traits are often defined by psychological models, with Big Five being the most popular one. In this section we briefly survey existing tools and datasets used for personality detection, and describe approaches to this task.

Importantly, we conclude that the existing annotated training datasets for personality classifier are \textbf{not sufficiently adequate} for the task, suffering from various drawbacks, as detailed below.

\subsection{Existing Personality Datasets}
\label{sec:existing-datasets}

Various datasets have been used for personality detection in prior studies. These datasets typically consist of textual data along with labeled personality traits (can be discrete or continuous), which are used to train and evaluate machine learning models. We describe the two most commonly used datasets for these models evaluation --- (1) Essays \citep{pennebaker1999linguistic} and (2) MyPersonality \citep{kosinski2015facebook} --- in more detail below.

\subsubsection{Essays Dataset}
Originating from 2,468 essays (collected between 1997 and 2004, each written by an undergraduate student), these writing submissions were part of a psychology course assignment \citep{pennebaker1999linguistic}. Students' personality scores were determined using the Big Five inventory, through a self-reported questionnaire that measures the five personality traits. Each essay was then labeled with five personality scores, based on a student's questionnaire, where the mean continuous scores where transformed into binary labels: "yes" (for high-), and "no" (for low-trait presence, respectively).

Despite its relative popularity, this dataset has several obvious limitations: First, annotating the entire essay (of average length of 663.10 tokens) with a single self-reported score is misleading, since different sections of the essay written by the same person may exhibit varying levels of each trait. Second, converting scores into binary labels with a median split oversimplifies the inherently continuous nature of a personality characteristic. Per our inspection, only a small ratio of texts can be associated with one of a trait extremes ("yes" or "no"), while the vast majority span the more neutral range. 

\subsubsection{MyPersonality Dataset}
MyPersonality dataset contains information from a Facebook application originally used by over 6 million users \citep{kosinski2015facebook}. Similarly to Essays, personality scores were assigned to authors based on self-reported personality questionnaire, along with several social network measures, including network size, density, brokerage, and transitivity. The final dataset includes textual Facebook status updates assigned with continuous scores for each personality trait per author. MyPersonality has been particularly useful for research because it combines social media text with personality trait labels, allowing the study of personality in a more naturalistic setting. However, the project was discontinued in 2018 due to the heavy burden of its maintenance:\footnote{\url{https://sites.google.com/michalkosinski.com/mypersonality}} its data is no longer publicly available, and only a minor subset of 9,917 texts by 250 users is available online.

MyPersonality dataset also has several notable drawbacks: First, the publicly available subset of the dataset is very limited, which significantly restricts the robustness of any analysis conducted. Furthermore, the self-reported personality scores are assigned to all user's statuses, regardless of how each trait manifests in individual texts, thereby posing a limitation, similar to that of Essays.

\begin{table}[h!]
\caption{Existing tools' accuracy on the two available datasets -- Essays and MyPersonality. Each text sample was judged with binary indication: low or high level for each of the Big Five traits. "*" indicates cases where the tool was documented to be pre-trained on (among others) the tested dataset. Best result per trait and dataset is boldfaced.}
\centering
\begin{tabular}{clrrrrr}
\toprule
test data & tool & OPN & CON & EXT & AGR & NEU \\
\midrule
\multirow{4}{*}{{\rotatebox[origin=c]{0}{Essays}}}
& (1) Psychology Insights       & \textbf{0.567} & 0.511 & \textbf{0.521} & 0.507 & 0.504 \\
& (3) Apply Magic Sauce         & 0.557 & 0.538 & 0.495 & 0.510 & \textbf{0.549} \\
& (4) Personality Prediction*   & 0.517 & 0.532 & 0.514 & 0.512 & 0.511 \\ \hdashline
& tools' majority vote          & 0.547 & \textbf{0.547} & 0.519 & \textbf{0.517} & 0.524 \\ \midrule
\multirow{4}{*}{{\rotatebox[origin=c]{0}{MyPersonality}}}
& (1) Psychology Insights*      & \textbf{0.844} & \textbf{0.688} & \textbf{0.652} & \textbf{0.608} & \textbf{0.812} \\
& (3) Apply Magic Sauce*        & 0.544 & 0.572 & 0.596 & 0.568 & 0.560 \\
& (4) Personality Prediction    & 0.684 & 0.448 & 0.424 & 0.524 & 0.396 \\ \hdashline
& tools' majority vote          & 0.780 & 0.612 & 0.624 & 0.596 & 0.644 \\
\bottomrule
\end{tabular}
\label{tbl:existing-tools-accuracy}
\end{table}

\subsection{Approaches to Personality Detection}
\label{sec:approaches-detection}

Prior to developing a novel dataset and model, we performed a thorough assessment of four freely available off-the-shelf tools: (1) {Psychology Insights},\footnote{\url{https://github.com/fuguixing/psychology-insights-frontend}} (2) {Personality Recognizer},\footnote{\url{https://farm2.user.srcf.net/research/personality/recognizer.html}} (3) {Apply Magic Sauce},\footnote{\url{https://applymagicsauce.com}} and (4) {Jkwieser’s Personality Prediction}.\footnote{\url{https://github.com/jkwieser/personality-prediction-from-text}} Notably, tools (1) and (3) were trained on the (undocumented) parts of the MyPersonality dataset, while tool's (4) training data includes Essays. Personality Recognizer (tool (2)) showed considerably inferior results; we excluded it from further experiments. Our choice of these tools was driven by their availability or the ease of reproducibility.

Evaluating performance of the three tools on both datasets yields disappointingly (but not surprisingly) low results. We report a tool's per-trait accuracy: the ratio of data examples assigned with the correct (binary) label out of the total amount. Considering the random baseline of 0.5, and despite the exposure of the tools to the datasets, the accuracy is low (inline with what is reported in prior art), questioning the applicability of these approaches to the task at hand. 

Table~\ref{tbl:existing-tools-accuracy} reports classification accuracy results of existing tools on the two available datasets. We attribute the poor results mainly to the limitations of the existing datasets (see Section~\ref{sec:existing-datasets}), and anticipate that \textit{high quality training data} coupled with contemporary modeling, will advance the state-of-the-art in this field. We describe our approach and methodology in the next section.

\section{Methodology}
\label{sec:methodology}

Considering the detailed drawbacks of the existing personality datasets, we decided to leverage the capabilities of generative AI for collecting a novel Big Five personality dataset that will be used to train personality classifiers. Contemporary LLMs excel at creating diverse, highly naturalistic textual content, indistinguishable from human writing \citep{wu2023survey, dathathri2024scalable}; we used {Google's Gemini}\footnote{\url{https://gemini.google.com/app}} and {OpenAI's GPT}\footnote{\url{https://chatgpt.com/}} for generating high-quality, diverse dataset(s) with short passages, each associated with low or high personality trait level. The five collected datasets (one per trait) were evaluated by human annotators, and used for training classification models for traits ranking on the continuous 0-1 scale. Below we provide details on dataset generation and model training.

\subsection{Big Five Generated Datasets (\texttt{GenBigFive})}
We applied a systematic approach to generate textual content that could be used to train and evaluate classification models for each of the Big Five personality traits, as described below.

\subsubsection{Definition Compilation}
We gathered detailed definitions for each of the Big Five personality traits (OPN, CON, EXT, AGR, NEU) from academic and online resources; these definitions were manually concatenated into a readable and comprehensive format to serve as the basis for our prompts.

\subsubsection{Prompt Creation}
For each personality trait, we wrote a primary prompt that started with the trait's definition and ended with a request for the LLM to generate several paragraphs demonstrating a high (and similarly, low) level of the trait. As a concrete example, for Extroversion, we generated texts "written by" people who are highly extroverted and then by those who score low on Extroversion, i.e. considered introverts. Appendix~\ref{sec:appendices} contains our final prompt definitions for the five traits.
%
We iteratively modified the requests to ensure the generated texts were unique, authentic, and avoided repetition. We also instructed the models to write as if they were different personas, iterating over a variety of topics and writer profiles, e.g., texts written by high school teenagers, by sports lovers (fans or practitioners), politicians, and also texts about finance, news, traveling, etc.

\subsubsection{Dataset Generation} Using GPT3.5 for training set collection, and Gemini for test set, we collected hundreds of short passages exhibiting the two extremes of each personality trait. 
Table~\ref{tbl:gen-big-five} reports the final statistics of the collected \texttt{GenBigFive} dataset, after cleanup and filtering, and Table~\ref{tbl:training-set-examples} presents a few generated text examples, produced as exhibiting "high" and "low" trait level by GPT3.5.

\begin{table}[h!]
\caption{Newly collected \texttt{GenBigFive} dataset statistics: GPT3.5 (further used for training) and Gemini (later used for test). The number of texts, split roughly by half between low and high trait level, and the mean number of words (WC) in a text are reported.}
\centering
\begin{tabular}{lcccc}
\toprule
& \multicolumn{2}{c}{GPT-3.5 (train)} & \multicolumn{2}{c}{Gemini (test)} \\
trait & texts & mean(WC) & texts & mean(WC) \\
\midrule
OPN & 272 & 62.25 & 144 & 63.13 \\
CON & 306 & 71.15 & 288 & 64.17 \\
EXT & 453 & 75.64 & 211 & 97.11 \\
AGR & 321 & 74.52 & 296 & 88.30 \\
NEU & 354 & 63.75 & 325 & 72.86 \\
\bottomrule
\end{tabular}
\label{tbl:gen-big-five}
\end{table}

\begin{table}[h!]
\caption{Example texts generated by GPT3.5 with high and low presence of two personality traits: EXT and AGR. The generated texts are purposefully exaggerated (by crafting prompts accordingly) towards high or low trait requirements, for training classifiers, which benefit from extreme examples.
}
\centering
\resizebox{\textwidth}{!}{
\begin{tabular}{cp{15cm}c}
\toprule
trait & text & rank \\
\midrule
EXT & Hey there! I absolutely thrive in social settings – the more people, the merrier! It's like every conversation is an opportunity to learn something new or share a laugh. I simply love being in the middle of it all, soaking up the energy and excitement around me. Whether it's a party, a networking event, or just a casual get-together, count me in! & high \\ \midrule
EXT & Social gatherings always leave me drained. It's not that I dislike people, but the constant buzz of conversation exhausts me. I find solace in quiet moments alone, where I can recharge my energy and gather my thoughts. [...] 
Being the center of attention is my worst nightmare; I much prefer blending into the background and observing rather than being in the spotlight. & low \\ \midrule \midrule
AGR & Protecting our planet and its delicate ecosystems has always been a priority close to my heart. Every action we take contributes to the health of our environment. Whether it's reducing our carbon footprint or advocating for renewable energy sources, we have the power to make a positive impact. & high \\
\midrule
AGR & It's utterly baffling how some people can't seem to handle the simplest tasks without constant hand-holding. I mean, come on, do I look like your personal assistant? Get it together and figure it out yourself for once. I've got my own stuff to deal with, and I certainly don't have time to babysit grown adults who can't take responsibility for themselves. & low \\
\bottomrule
\end{tabular}
}
\label{tbl:training-set-examples}
\end{table}

\subsubsection{Human Evaluation} 
We evaluated the quality of generated data through human annotation of 250 samples: 50 samples were randomly selected for each trait, split equally between low and high. We used the {Appen} platform\footnote{\url{https://www.appen.com/}} to recruit native English speakers, who were presented with a personality trait description (identical to what the LLM was prompted with, see Appendix~\ref{sec:guidelines}) along with a few examples. Quizzes were integrated into the task to ensure high-quality outcome. Each sample was annotated by five annotators, with an exception of 17 samples, that were annotated only by three. Each example was assigned a trait level (low or high) based on the majority vote.

With sufficient training, the task turned out to be relatively easy: annotation agreement with GPT3.5 generation was 94.7\% and inter-annotator agreement was $k${=}0.76.

\subsection{Building New Personality Classifier(s)}
\label{sec:building-classifier}

We used the collected datasets for each of the five traits for training reliable and accurate personality classifiers. Here, we follow the approach successfully used in prior studies for classification of \textit{emotion} \citep{aggarwal2020exploration} and the degree of \textit{concept abstractness} \citep{francis2021quantifying}, training a logistic regression classifier with passage embeddings as feature vectors. This methodology was chosen for its simplicity, effectiveness, the inherent ability to produce a continuous output score (posterior probability), and efficiency (fast training and inference). 

We trained five logistic regression models, one per trait, where the passages with low and high trait level were encoded into embeddings using the \texttt{intfloat/e5-large-v2} encoder \citep{wang2022text},\footnote{\url{https://huggingface.co/intfloat/e5-large-v2}} due to its proven benefits on multiple tasks.
We trained the models on GPT3.5-generated data (train set) and tested on Gemini-generated data (test set); that, in order to avoid possible confounds in train and test data, generated by the same LLM. We also tested the off-the-shelf tools (see Section~\ref{sec:approaches-detection}), showing that the newly trained classifiers yield superior results on the unseen test set.

\subsubsection{Evaluation Results}
We report evaluation accuracy of the five classifiers in Table~\ref{tbl:lr-evaluation-results}. As a complementary experiment, we also trained the classifiers on the GPT3.5-generated dataset combined with the MyPersonality dataset, and tested on the Gemini-generated test set. High test set accuracy is achieved for each of the five traits, with Openness (OPN) obtaining the lowest accuracy of 0.874. Additional MyPersonality training data harms the results in all cases but CON. Finally, existing tools yielded inferior, nearly random results.

\begin{table}[h!]
\caption{Our classifiers' accuracy, when predicting the Big Five personality traits on the unseen Gemini-generated test set. We report results for two settings: (1) train set comprising solely the GPT3.5-generated data, and (2) + the MyPersonality dataset. Best result in a column is boldfaced.
}
\centering
\begin{tabular}{lrrrrr}
\toprule
train set & OPN & CON & EXT & AGR & NEU \\
\midrule
GPT3.5      & \textbf{0.874} & 0.934 & \textbf{0.963} & \textbf{0.959} & \textbf{0.991} \\
GPT3.5+MP   & 0.526 & \textbf{0.954} & 0.848 & 0.939 & 0.731 \\
\bottomrule
\end{tabular}
\label{tbl:lr-evaluation-results}
\end{table}

We conclude that the logistic regression classifier trained on the newly-created \texttt{GenBigFive} dataset can be used to reliably label our music-fans posts and comments for personality traits.

\section{Experimental Results and Discussion}
\label{sec:experiments}

We next applied the \texttt{intfloat/e5-large-v2} encoder (see Section~\ref{sec:building-classifier}) to over 500K Reddit posts and comments, written by nearly 5,000 Redditors with five distinct musical preferences, roughly 1,000 for each genre. We used the pre-trained classification models for prediction, assigning each text with five continuous scores, across five personality trait dimensions. 

Table~\ref{tbl:user-text-traits-examples} presents example posts from our dataset, along with their automatically assigned low and high personality scores.

\begin{table}[h!]
\caption{Example user texts (verbatim from the dataset) illustrating varying levels of the Openness (OPN) and Neuroticism (NEU) traits, along with a probability score indicating the likelihood of high trait presence (posterior), as produced by the respective regression models. 
}
\centering
\resizebox{\textwidth}{!}{
\begin{tabular}{cp{15cm}r}
\toprule
trait & text (taken verbatim from the Reddit dataset) & score \\
\midrule
OPN & Wow....there is so much awesome in this artwork. Pretty amazing {\textless{redditor username}\textgreater} could render such distinct subjects as the numerous blooming flowers, the Greek vase fragmented story and the bronze relief sculpture. Mind blowing visually. I bet there's language symbolized in the flowers interacting with the decorative elements. & 0.754 \\ \midrule
OPN & Yup you pretty much hit the nail on the head for me. My partner of 2 years is pretty vocal about marriage and her parents have been married since before she was born. My parents split when I was super young so I pretty much haven't put too much thought into marriage into my relationship. I just figured I'll see how I feel in the moment when our relationship hits 5y or something... & 0.254 \\ \midrule \midrule
NEU & feel like i'm going crazy. everything is up in the air. no idea where i'll be in august. i'm applying for jobs and my soul tears apart each time because the process is designed to be fucking excruciating and make you feel worthless (so they can hire you cheaper). [...] & 0.811 \\ \midrule
NEU & This is great. Eve and Amanda and Elza Brabant talk about the 3 reasons we eat: we eat for nutrients, we eat for pleasure, and we eat for community. I like how you are thinking: I think what you are doing enhances your eating. Best. & 0.184 \\
\bottomrule
\end{tabular}
}
\label{tbl:user-text-traits-examples}
\end{table}

\subsection{Human Evaluation}
We conducted human evaluation, this time for labeling actual Redditors' content, scored by the classifiers, to make sure the predictive ranking fits human intuition. Similarly to the first task, 250 samples in total where selected for annotation --- where the continuous score was binarized into two extremes --- this time yielding lower inter-annotator agreement of $k$={0.67} and agreement of 86.2\% with scores (low or high) assigned by the models. We note that the lower agreement (recall the 94.7\% for generated content) is expected since human annotation is compared to the (imperfect) classifiers outcome.

\subsection{Authors' Personality Exhibits Association with their Musical Preferences}
Next we assigned each user with scores, reflecting their mean personality ranking on each of the Big Five traits. Specifically, we averaged the prediction scores across all texts authored by a given user. Consequently, a community-level personality trait score was computed by averaging over individual scores of users in the specific community. As a concrete example, the mean OPN score of the Classical music lovers was calculated by averaging over the OPN score of the 982 this genre's fans in our dataset.

\begin{figure}[h!]
\centering
\resizebox{1.0\linewidth}{!}{
\includegraphics{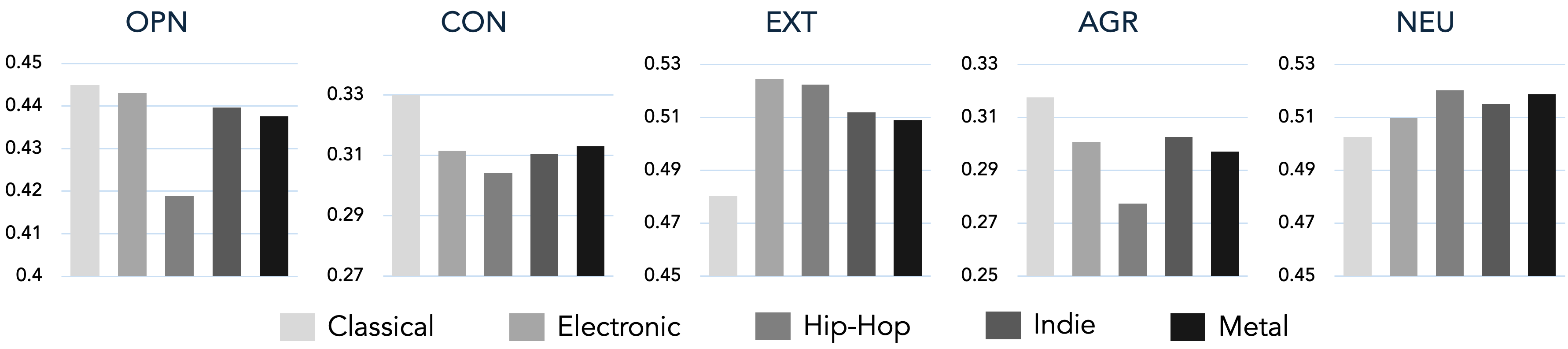}
}
\caption{Mean personality trait by musical genre fans. All five traits exhibit significant within-community differences, as measured by the ANOVA test.}
\label{fig:main-results}
\end{figure}

Figure~\ref{fig:main-results} presents the results. Many of the findings align with common intuition and prior results in non-linguistic studies \citep{anderson2021just, greenberg2022universals}: Classical music enthusiasts tend to exhibit the highest levels of Agreeableness, while Hip-Hop fans show the lowest levels, compared to all other genres. Additionally, Classical music enthusiasts generally show lower levels of Extroversion, Hip-Hop and Metal fans tend to have higher levels of Neuroticism. 

\subsection{Statistical Significance of the Findings}
We turn to evaluate the significance of differences between the groups' scores. Our primary statistical test is ANOVA \citep{girden1992anova} -- a statistical method that compares means across multiple groups to identify statistically significant differences. It analyzes both within-group and between-group variation to determine if observed differences are due to actual effects or random variation. Particularly useful for experiments involving more than two groups, ANOVA extends the capabilities of the two-group t-test. 
ANOVA's primary output, the F-statistic, indicates whether the means of the groups are significantly different. A $p$-value lower than 0.05 is typically used as significance threshold.

When applied on our per-community personality scores across each dimension, the test results --- all virtually zero, significant at {$p${<}5.0e-46} --- strongly suggest that there are considerable differences in the means across the groups for each personality trait; the extremely low $p$-values confirm that these differences are significant. 

\subsubsection{Pairwise Significance and Effect Size}
While ANOVA provides a broad analysis by testing whether there are any significant differences across multiple groups, it does not specify where those differences lie. We further applied pairwise group analysis using two-tailed unpaired test for difference in means; specifically, we used the non-parametric Mann-Whitney test \citep{mann1947test}, that does not imply assumptions on the underlying population distribution.

Statistical significance tests benefit from large samples under test, where even extremely small differences may show up significant. A common way to overcome this bias, is by reporting the additional measure of \texttt{effect-size}. As such, Cohen's $d$ effect size \citep{cohen2013statistical} quantifies the magnitude of the difference between two groups in terms of standard deviations, helping to understand the practical significance of the observed differences. A higher Cohen's~$d$ absolute value indicates a larger difference between the groups, where the common interpretation is as follows: absolute Cohen's $d$ value between 0.2 and 0.5, denotes a small effect size, 0.5--0.8 -- medium, and 0.8 or higher indicates high effect size.

\begin{figure}[h!]
\centering
\resizebox{0.70\textwidth}{!}{
\includegraphics{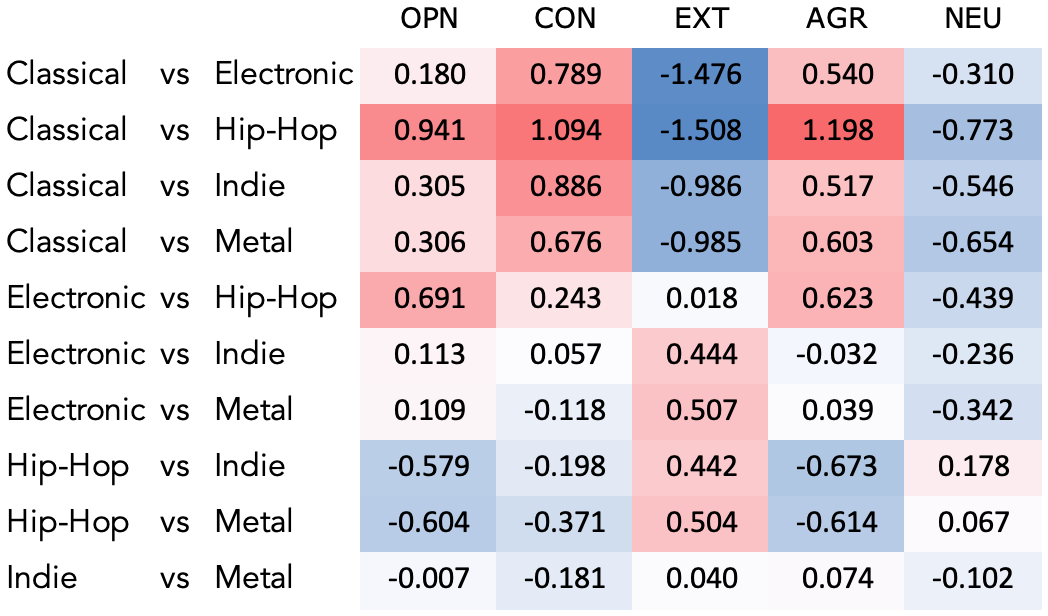}
}
\caption{Cohen's $d$ effect size and pairwise Mann-Whitney test. The value in each cell indicates the effect size, its color -- whether the difference is significant at the $p$-val<0.05 level. Positive effect size (left group has a higher trait value than right) is marked with red, while negative effect size -- with blue; both denote significant difference. Uncolored cells represent insignificant difference.}
\label{fig:cohensd-heatmap}
\end{figure}

Figure~\ref{fig:cohensd-heatmap} presents our pairwise findings. Evidently, there exist significant associations between musical preferences and personality traits. Classical music listeners are notably less extroverted compared to both Hip-Hop and Electronic fans, as indicated by large negative effect sizes. Additionally, Classical music lovers exhibit higher levels of Agreeableness, particularly when compared to Hip-Hop listeners. In terms of Openness, Hip-Hop shows the lowest presence compared to the other genres, especially Classical and Electronic. Also, Hip-hop and Metal listeners display higher levels of Neuroticism when compared to Classical music fans. Our findings suggest that the degree of certain trait, as detected in authentic writing, is strongly associated with a person's musical preferences.

\subsection{Predicting Genre from Personality}
As an additional experiment, we approach the challenging task of predicting a user's musical preference based solely on a five-dimensional personality vector derived from their authentic writing. Specifically, we trained another logistic regression classifier for five-class classification task: given a personality feature vector with five scores --- OPN, AGR, CON, EXT, NEU --- predict the individual's musical preference out of the five genres we consider in this study. Notably, each user is associated in our study with a single genre (see Section~\ref{sec:dataset}). Applied on a random split of the 4,683 users into 80-20 train and test, the classifier achieved the accuracy of 0.424, considerably exceeding the five-class random baseline of 0.2: while insufficient for any practical purpose, this finding supports once again that personality traits "shine through" our language, and here, as well, are associated with our musical preferences.

\section{Conclusions}
\label{sec:conclusions}

Our study presents a comprehensive analysis of the relationship between musical preferences and personality traits as those are extracted from textual data from social media. Using a carefully curated dataset and reliably trained models, we reveal significant associations between personality traits, as defined by the Big Five model and extracted from text, and five distinct musical genres. Ultimately, this research opens up new possibilities for understanding how our personalities shape, and are shaped by, the music we engage with, offering a compelling intersection of language, music and psychology. Additionally, we release resources for future research.

\section*{Ethical Considerations}
\label{sec:ethical}

We use publicly available data to study how individuals' musical preferences are reflected in their spontaneous language through the lens of the Big Five personality traits. The use of publicly available data from social media platforms, such as Reddit, may raise normative and ethical concerns. These concerns are extensively studied by the research community as reported in e.g., \citet{proferes2021studying}. 

Here we address two main concerns: \textbf{Anonymity:} Data used for this research can only be associated with participants' user IDs, which, in turn, cannot be linked to any identifiable information, or used to infer any demographic trait. 
\textbf{Consent:} \citet{jagfeld2021understanding} debated the need to obtain informed consent for using social media data mainly because it is not straightforward to determine if posts pertain to a public or private context. Ethical guidelines for social media research \citep{benton2017ethical} and practice in comparable research projects \citep{ahmed2017using}, as well as {Reddit's terms of use},\footnote{\url{https://www.redditinc.com/policies/user-agreement-september-12-2021}} regard it as acceptable to waive explicit consent if users' anonymity is protected.

We hired human annotators (native English speakers from specific geographies) for annotating samples of data for the presence of personality traits during this study. The annotators were hired via the {Appen} annotation platform,\footnote{\url{https://www.appen.com/}} and were payed above the US federal minimum wage. Comments left by our annotators at the end of the task indicate they found the work interesting and enjoyable.
\section*{Limitations}
\label{sec:limitations}

Despite our efforts to carefully mitigate potential confounds, this study is not without limitations. The use of Reddit as a data source introduces certain biases, as its demographic skews toward younger males, which may affect the generalizability of our findings. Additionally, assuming that the most active participants in genre-specific subreddits are representative of the broader fan base is a strong, though not unreasonable, assumption, as not all subscribers may strongly identify with the genre. Moreover, (unknown) factors such as age and gender were not explicitly controlled for, potentially influencing the results.
Nevertheless, we believe that the large scale of our dataset helps enhance generalizability and mitigate demographic biases. Furthermore, prior research has validated the approach of using high activity in topic-related subreddits as a proxy for user affiliation with that topic.

\begin{acknowledgments}
We are grateful to our four anonymous reviewers for their useful comments and constructive feedback.
\end{acknowledgments}

\section*{Declaration on Generative AI}
The authors have not employed any Generative AI tools while writing the paper.

\bibliography{custom}

\appendix
\section{Appendices}
\label{sec:appendices}

\subsection{Personality Traits (Short) Definitions}

\paragraph{Openness} 
Openness (also referred to as Openness to experience) emphasizes imagination and insight the most out of all five personality traits. People who are high in Openness tend to have a broad range of interests. They are curious about the world and other people and are eager to learn new things and enjoy new experiences. People who are high in this personality trait also tend to be more adventurous and creative. Conversely, people low in this personality trait are often much more traditional and may struggle with abstract thinking. Intellect, imagination, and Openness describe your imagination and how creative you are. It refers to your sense of curiosity about the world and your willingness to try new things, and to be exposed to new experiences.

\paragraph{Conscientiousness}
Conscientiousness is a trait that refers to how thoughtful and goal-oriented you are. It reflects the degree of your control over your impulses and your level of organization and work ethic. Conscientiousness describes a person's ability to regulate impulse control in order to engage in goal-directed behaviors. It measures elements such as control and persistence of behavior.

\paragraph{Extroversion}
Extroversion (or extraversion) is a personality trait characterized by excitability, sociability, talkativeness, assertiveness, and high amounts of emotional expressiveness. People high in extroversion are outgoing and tend to gain energy in social situations. Being around others helps them feel energized and excited. People who are low in this personality trait or introverted tend to be more reserved. They have less energy to expend in social settings and social events can feel draining. Introverts often require a period of solitude and quiet in order to "recharge". Extroversion reflects how you interact socially. It describes your emotional expression and how comfortable you are in your environment.

\paragraph{Agreeableness}
This personality trait includes attributes such as trust, altruism, kindness, affection, and other prosocial behaviors. People who are high in agreeableness tend to be more cooperative while those low in this personality trait tend to be more competitive and sometimes even manipulative. Agreeableness is a personality trait that describes how you treat your relationships with others. It reflects how kind and helpful you are toward people. Overall, high agreeableness means you desire to keep things running smoothly and value social harmony.

\paragraph{Neuroticism}
Neuroticism is a personality trait characterized by sadness, moodiness, and emotional instability. Individuals who are high in neuroticism tend to experience mood swings, anxiety, irritability, and sadness. People low in this trait tend to be more stable and emotionally resilient. Neuroticism is a personality trait that refers to your emotional stability. As a personality dimension, neuroticism is characterized by unsettling thoughts and feelings of sadness or moodiness.

\subsection{Guidelines for Human Annotators (used also as Prompts for Generative Models)}
\label{sec:guidelines}

For each personality trait, we wrote a primary prompt that started with the trait's definition and ended with a request for the LLM to generate paragraphs demonstrating a high (and similarly, low) level of the trait. As a concrete example, for Extroversion, we generated texts "written by" people who are highly extroverted and then by those who rank low on Extroversion, i.e. considered introverts.

We iteratively modified the requests to ensure the generated texts were unique, authentic, and avoided repetition. We also instructed the models to write as if they were different personas, iterating over a variety of topics and writer profiles, e.g., texts written by high school teenagers, by sports lovers (fans or practitioners), politicians, and also texts about finance, news, traveling, etc. The models were presented with multiple (found in literature) examples for texts with high and low personality trait level.\footnote{All five traits guidelines and prompts are available in the data that will be released upon this paper acceptance.}

\subsubsection{Prompt Used for Collecting Openness Texts}
We used the following prompt (also used as a trait description to human annotators:)

\paragraph{Openness Trait Description:} Openness (also referred to as Openness to experience) emphasizes imagination and insight the most out of all five personality traits. People who are high in Openness tend to have a broad range of interests. They are curious about the world and other people and are eager to learn new things and enjoy new experiences. People who are high in this personality trait also tend to be more adventurous and creative. Conversely, people low in this personality trait are often much more traditional and may struggle with abstract thinking. Intellect, imagination, and Openness describe also how creative you are. 

Openness to experience refers to one's willingness to try new things as well as engage in imaginative and intellectual activities. It includes the ability to "think outside of the box", curiosity about and tolerance for diverse cultural and intellectual experiences.

People who are considered to have high Openness are more likely to be: Very creative, Open to trying new things, Focused on tackling new challenges, happy to think about abstract concepts, enjoy learning and trying new things, have an active imagination, be more creative, be intellectually curious, think about abstract concepts, enjoy challenges, like to travel, have a wide range of interests, Curious, Imaginative, Creative, Open to trying new things, Unconventional. They have a basic tendency for Actions (a need for variety, novelty, and change). Interest in travel, many different hobbies, knowledge of foreign cuisine, diverse vocational interests, and friends who share tastes.
People who are considered to have low Openness are more likely to Dislike change, Do not enjoy new things, Resist new ideas, Not very imaginative, Dislike abstract or theoretical concepts, dislike change, be likely to stick to routines, not be imaginative or creative, have more traditional thinking, be more grounded, Predictable, Not very imaginative, Dislikes change, Prefer routine, Traditional.

Openness vs. Closedness to Experience: Those who score high on Openness to experience are perceived as creative and artistic. They prefer variety and value independence. They are curious about their surroundings and enjoy traveling and learning new things. People who score low on Openness to experience prefer routine. They are uncomfortable with change and trying new things, so they prefer the familiar over the unknown. As they are practical people, they often find it difficult to think creatively or abstractly.

\textbf{<for models> Consider the following task:} Please provide 10 paragraphs of 40-150 words each, "written by" people with high Openness personality. All the paragraphs should be very diverse and should not be repeated at all. Don't even repeat sentences. Let's start with the first paragraph and then continue with 9 more iterations. Pretend to be different males or females of a variety of ages, with different socioeconomic statuses with high Openness, i.e. the Openness personality trait is strong and well noticeable in their texts. Avoid repeating the word Openness in your writing. You can tell personal details about the writer but don't introduce them with details in a formal manner (don't start with name, age and occupation!). The texts should not be about the writer but written by them.

\paragraph{Alternative Endings}
Several alternative closing instructions were further provided in the prompt for generation of diverse content:

(a) Write distinct 40-150-word length paragraphs. The paragraphs should be written by totally different people, but all should have in common the strong Openness personality trait. Try to make the paragraphs unique and avoid repeating yourself.

(b) Generate 10 paragraphs written by characters with low Openness. Here are the rules: 1. The paragraph length should be 40 to 150 words. 2. Topics should be random and not repeated. 3. Avoid repeating sentences. 4. Use casual daily internet language. 5. Texts should not be about Openness at all but demonstrate low Openness. 6. Don't introduce the character at the beginning.

(c) Generate 10 distinct paragraphs, each should be 40-150 words long, about different random topics. Paragraphs or sentences (or even parts of sentences) should not be repeated as much as possible. The texts should be "written by" characters with high Openness. Use causal daily language and don't mention the term Openness.

\paragraph{<for annotators> Annotate Passages for their Author's Openness}
You are presented with a passage written by a person (or generated by a model). Read the passage \textbf{carefully} and decide if its author is likely to be characterized by low or high level of the Openness personality trait. Use a binary rank ("low" or "high") even in cases where the decision is not a clear-cut, based on the likelihood of the writer to have low or high trait presence.


\end{document}